\documentclass[letterpaper, 10 pt, conference]{ieeeconf}  % Comment this line out if you need a4paper

\IEEEoverridecommandlockouts                              % This command is only needed if 
\makeatletter
\let\NAT@parse\undefined
\makeatother

\usepackage[utf8]{inputenc} % Allow utf-8 input
\usepackage[T1]{fontenc}    % Use 8-bit T1 fonts
\usepackage{textcomp}       % Provides extra text symbols
\usepackage{nicefrac}       % Compact symbols for 1/2, etc.
\usepackage{placeins}       % For \FloatBarrier
\usepackage{microtype}      % Microtypography for better spacing

\usepackage{amsmath,amssymb,amsfonts} % AMS math packages

\usepackage{graphicx}       % For including images
\graphicspath{{media/}}     % Set image folder
\usepackage{subcaption}     % For subfigures (e.g., Fig 1a, 1b)
\usepackage{booktabs}       % Professional-quality tables
\usepackage{multirow}       % For multi-row cells in tables
\usepackage{adjustbox}      % To adjust box content (like tables)
\usepackage{pdflscape}      % For landscape pages
\usepackage{longtable}      % For tables that span multiple pages
\usepackage{stfloats}       % Better control over float placement
\usepackage{float}          % Finer control over float placement

\usepackage{algorithm}
\usepackage{algpseudocode}

\usepackage{cite}           % Handlessubcitations correctly for IEEE
\usepackage{url}            % Simple URL typesetting
\usepackage{hyperref}       % Hyperlinks (should be loaded late)
\usepackage{orcidlink}      % For ORCID iDs

\usepackage{fancyhdr}       % For custom headers/footers
\usepackage{xcolor}         % For using colors
\usepackage{lipsum}         % Dummy text for layout testing

\definecolor{darkblue}{rgb}{0.0, 0.0, 0.55}
\hypersetup{
    colorlinks=true,
    linkcolor=darkblue,
    citecolor=darkblue,
    filecolor=magenta,      
    urlcolor=black,
}

\title{\LARGE \bf
A Synthetic Dataset for Manometry Recognition in Robotic Applications}

\author{Pedro Antonio Rabelo Saraiva$^{1}$, Enzo Ferreira de Souza$^{1}$, João Manoel Herrera Pinheiro$^{1}$,\\ Thiago H. Segreto$^{1}$, Ricardo V. Godoy$^{1}$, and Marcelo Becker$^{1}$% <-this % stops a space
\thanks{This work was carried out with the support of Petrobras, using resources from the R\&D clause of the ANP, in partnership with the University of São Paulo (USP) and the intervening foundation Fundação de Apoio à Física e à Química (FAFQ), under Cooperation Agreement No. 2023/00013-7 and 2023/00016-6.}
\thanks{$^{1}$Pedro Saraiva, Enzo Ferreira de Souza, Thiago H. Segreto, Ricardo V. Godoy, and Marcelo Becker are with the  Department of Mechanical Engineering, University of São Paulo, São Carlos, Brazil.
{\tt\small becker@sc.usp.br}}%}%
}

\begin{document}
\maketitle
\thispagestyle{empty}
\pagestyle{empty}

%%%%%%%%%%%%%%%%%%%%%%%%%%%%%%%%%%%%%%%%%%%%%%%%%%%%%%%%%%%%%%%%%%%%%%%%%%%%%%%%
\begin{abstract}
This paper addresses the challenges of data scarcity and high acquisition costs in training robust object detection models for complex industrial environments, such as offshore oil platforms. Data collection in these hazardous settings often limits the development of autonomous inspection systems. To mitigate this issue, we propose a hybrid data synthesis pipeline that integrates procedural rendering and AI-driven video generation. The approach uses BlenderProc to produce photorealistic images with domain randomization and NVIDIA’s Cosmos-Predict2 to generate physically consistent video sequences with temporal variation. A YOLO-based detector trained on a composite dataset, combining real and synthetic data, outperformed models trained solely on real images. A 1:1 ratio between real and synthetic samples achieved the highest accuracy. The results demonstrate that synthetic data generation is a viable, cost-effective, and safe strategy for developing reliable perception systems in safety-critical and resource-constrained industrial applications.
% This work addresses the challenges of data scarcity and high acquisition costs for training robust object detection models in complex industrial environments, such as offshore oil platforms. The practical and economic barriers to collecting real-world data in these hazardous settings often hamper the development of autonomous inspection systems. To overcome this, in this work we propose and validate a hybrid data synthesis pipeline that combines procedural rendering with AI-driven video generation. Our methodology leverages BlenderProc to create photorealistic images with precise annotations and controlled domain randomization, and integrates NVIDIA's Cosmos-Predict2 world-foundation model to synthesize physically plausible video sequences with temporal diversity, capturing rare viewpoints and adverse conditions. We demonstrate that a YOLO-based detection network trained on a composite dataset, blending real images with our synthetic data, achieves superior performance compared to models trained exclusively on real-world data. Notably, a 1:1 mixture of real and synthetic data yielded the highest accuracy, surpassing the real-only baseline. These findings highlight the viability of a synthetic-first approach as an efficient, cost-effective, and safe alternative for developing reliable perception systems in safety-critical and resource-constrained industrial applications.

\end{abstract}

%%%%%%%%%%%%%%%%%%%%%%%%%%%%%%%%%%%%%%%%%%%%%%%%%%%%%%%%%%%%%%%%%%%%%%%%%%%%%%%%
\section{INTRODUCTION}

The uninterrupted functioning and security of vital infrastructure, especially within the petroleum and natural gas sectors, are of paramount importance due to the inherent dangers associated with the handled materials and the severe repercussions of system malfunctions~\cite{https://doi.org/10.1002/prs.10010,https://doi.org/10.1002/prs.11829,10.1007/978-3-319-67361-5_37}. Conventional maintenance and inspection procedures in these intricate industrial settings are largely dependent on human personnel, who are frequently exposed to substantial hazards such as contact with toxic agents, extreme temperatures, and confined spaces~\cite{KHAN1999361,YANG2023105061}. Accordingly, there is a strong demand for technologies that increase inspection effectiveness~\cite{Khan2022,s25154873} while simultaneously reducing human exposure in dangerous scenarios~\cite{robotics10020067,9290132,10758775}.

The introduction of autonomous legged robots, e.g., Unitree B2~\cite{robotics14050057} and ANYmal C~\cite{Hutter02092017}, offers a promising alternative~\cite{Trevelyan2016,9106415}. Unlike wheeled or tracked systems that struggle with uneven terrain, legged platforms can operate in a reliable manner even in the presence of stairs and irregular surfaces~\cite{ramezani2020legged}, enabling routine, automated data capture that minimizes downtime and reduces human risk~\cite{10.1007/978-981-15-9460-1_18}.

However, reliable perception remains a key bottleneck: robots must detect and localize facility-critical assets—valves, pipes, gauges, and safety equipment—to assess conditions and comply with procedures in visually cluttered plants~\cite{s21051571,10.1007/978-3-319-67361-5_37}. Acquiring and annotating large, diverse datasets for training modern detectors (e.g., YOLO) is time-consuming, costly, and often disruptive to operations; access constraints and safety risks further limit in-plant data collection~\cite{technologies12020015,10.1186/s40537-019-0197-0,redmon2016lookonceunifiedrealtime,bochkovskiy2020yolov4optimalspeedaccuracy}.

To overcome these limitations, in this work, we leverage high-quality synthetic data at two complementary levels. First, we use BlenderProc—an open-source, procedural rendering pipeline—to generate photorealistic images with controlled domain randomization over geometry, illumination, materials, and sensors~\cite{denninger2019blenderproc,s21237901,s23063013}. Second, we integrate NVIDIA Cosmos-Predict2, NVIDIA’s latest World Foundation Model (WFM) family for “physical AI”~\cite{nvidia_cosmos_platform,cosmos_predict2_research,cosmos_predict2_github}, via ComfyUI~\cite{comfyui_cosmos_predict2} workflows to synthesize physically plausible imagery and video priors from rare viewpoints, adverse lighting, and hard-to-stage events, thus broadening coverage beyond what is practical to render or capture in the field. Cosmos provides model variants and deployment tooling intended for robotics and autonomous vehicles (AV) pipelines, enabling scalable generation and curation that complements procedural graphics and reduces the sim-to-real gap in downstream perception~\cite{nvidia_cosmos_docs}.

This work introduces a practical hybrid synthetic plus real pipeline. Blending BlenderProc rendering with Cosmos-Predict2 generation for industrial asset perception, we report a case study on analog pressure gauges (manometers) and show that the same pipeline readily transfers to valves, digital or analog gauges and displays, flowmeters, placards, tags, and other safety-critical industrial devices with minimal changes, primarily swapping the 3D models. We build a large-scale, richly annotated synthetic dataset and train a YOLO-based detector, showing that synthetic data can substantially reduce reliance on costly and hazardous in-plant collection while improving recognition of safety-critical assets in complex environments.

The rest of this paper is organized as follows. Section~\ref{sec:rw} presents the related works. Section~\ref{sec:metho} details the unified pipeline procedural dataset creation with BlenderProc and AI-extended synthesis with Cosmos-Predict2 (ComfyUI) and training/validation of the object detector. Section~\ref{sec:results} presents quantitative and qualitative results, while Section~\ref{sec:conc} concludes with limitations and future work.

\section{Related Work}\label{sec:rw}

Synthetic data has become a practical solution for the scarcity and safety constraints of in-plant data collection, with domain randomization and photorealistic rendering shown to reduce the sim-to-real gap for perception and control~\cite{Paulin2023}. Early sim-to-real studies demonstrated that heavy appearance randomization can transfer object localization and even collision-avoidance policies trained purely in simulation to the real world~\cite{tobin2017domain,sadeghi2016cad2rl}. Recent analyses formalize when and how randomization narrows the gap and how to tune distributions over simulator parameters~\cite{vuong2019dr,chen2021udr}.

On the procedural/graphics side, toolchains such as BlenderProc provide modular, scriptable generation of photorealistic images with rich annotations and sensor simulation—now a common backbone for dataset bootstrapping in robotics and industrial vision~\cite{denninger2019blenderproc}. Parallel efforts in commercial-grade simulators (NVIDIA Omniverse/Isaac Sim Replicator) expose APIs for large-scale synthetic data generation with domain randomization, sensor models, and annotators, frequently used in AV/robotics pipelines~\cite{omniverse_replicator_docs, isaacsim_replicator_docs}. Unity’s Perception stack similarly targets turnkey dataset creation with labelers and randomized scenes for object detection and pose estimation~\cite{unity_perception_arxiv}.

Complementary to procedural rendering, world foundation models (WFMs) have emerged as AI-native generators that simulate and predict physically plausible scenes directly from text/visual conditions. NVIDIA’s Cosmos platform targets “physical AI” use cases (robotics, AV, industrial vision), exposing Cosmos-Predict2 models for Video-to-World generation (2B and 14B variants) and post-training/fine-tuning workflows aimed at scalable coverage of rare viewpoints, adverse illumination, and hard-to-stage events~\cite{nvidia_cosmos_site,nvidia_cosmos_docs_predict2,nvidia_cosmos_quickstart}. Public releases and reports emphasize open accessibility and training on large corpora of robotics/drive videos, positioning Cosmos as a data engine to complement graphics-based pipelines rather than replace them~\cite{nvidia_cosmos_blog_open,world_foundation_models_blog}. Community workflows via ComfyUI further reduce integration friction, enabling mixed procedural and generative curation where Cosmos assets are inserted into Blender/YOLO pipelines~\cite{comfyui_cosmos_examples}. Taken together, recent work suggests that hybrid pipelines, which incorporate procedural rendering for precise labels and controllable physics, along with WFM-based synthesis for broader coverage and temporal variation, can significantly reduce reliance on hazardous and costly field acquisition in complex industrial environments.

\begin{figure*}[!b]
    \centering
    \includegraphics[width=0.8\textwidth]{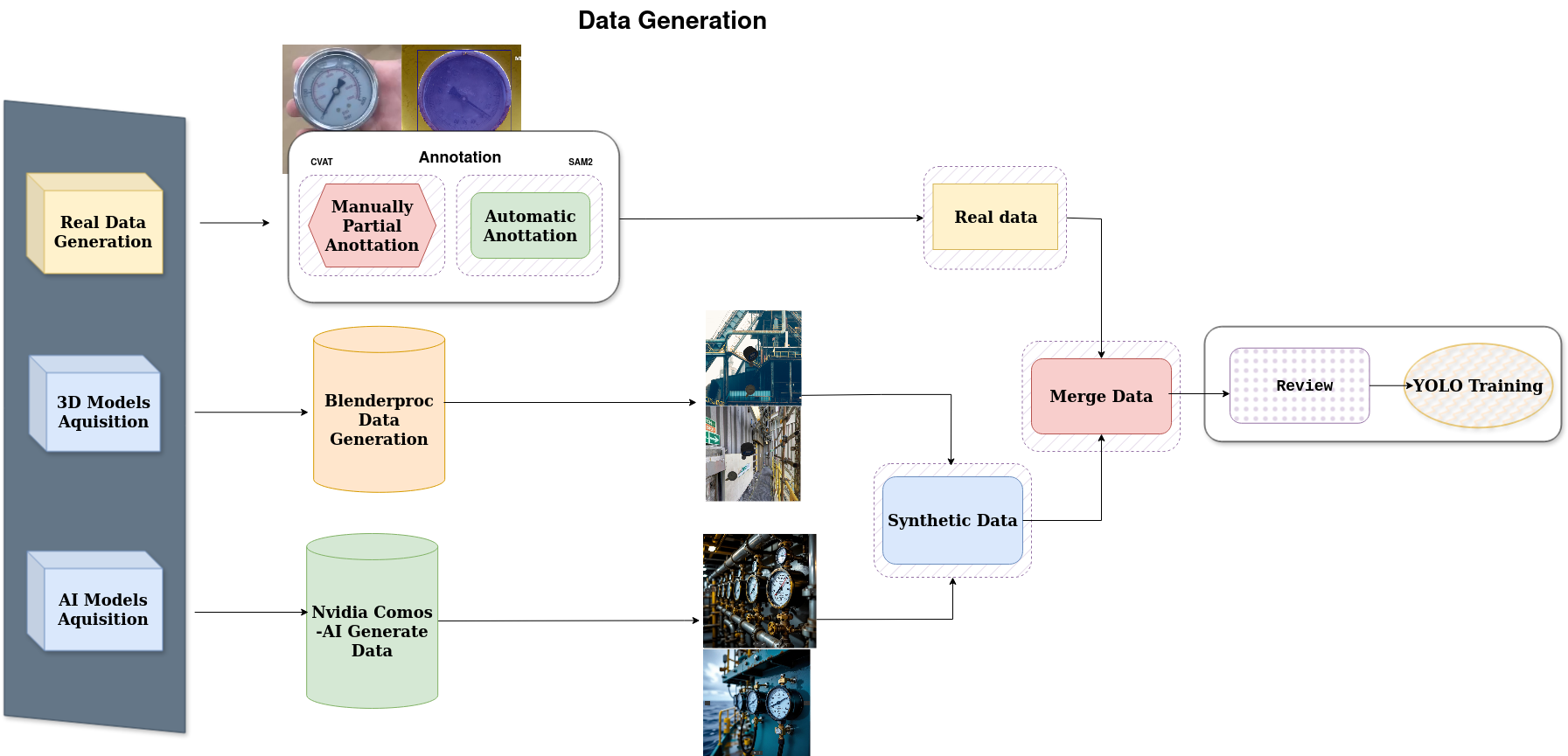}
    \caption{Proposed methodology pipeline for dataset creation and training. 
    The framework integrates three sources: (i) real data acquisition and 
    annotation (manual via CVAT \textit{(Computer Vision Annotation Tool)} and semi-automatic via SAM2 \textit{(Segment Anything Model 2)}), 
    (ii) synthetic renders generated in BlenderProc with domain randomization 
    (backgrounds, illumination, pose, post-processing), and 
    (iii) AI-extended video frames generated through ComfyUI with 
    NVIDIA Cosmos-Predict2 (relighting, motion, occlusion). 
    All data sources are reviewed, merged, and used for training the YOLO-based 
    object detector.}
    \label{fig:pipeline_unificado}
\end{figure*}

\section{Methodology}\label{sec:metho}

Our methodology combines real data acquisition, synthetic data generation
through BlenderProc, and AI-extended video synthesis using NVIDIA Cosmos-Predict2 integrated via ComfyUI. The unified pipeline is illustrated in
Fig.~\ref{fig:pipeline_unificado}.

\subsection{Real Data Acquisition and Annotation}
\begin{figure}[H]
  \centering
  \begin{subfigure}[b]{0.45\linewidth}
    \centering
    \includegraphics[width=\linewidth]{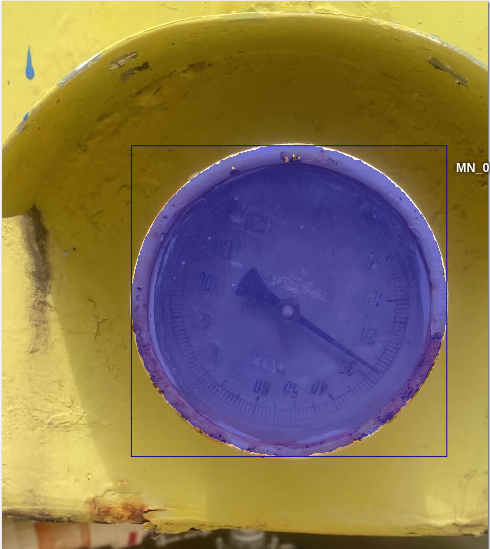}
    \caption{Manual bounding boxes in CVAT.}
    \label{fig:real_cvat_bbox1}
  \end{subfigure}\hfill
  \begin{subfigure}[b]{0.48\linewidth}
    \centering
    \includegraphics[width=\linewidth]{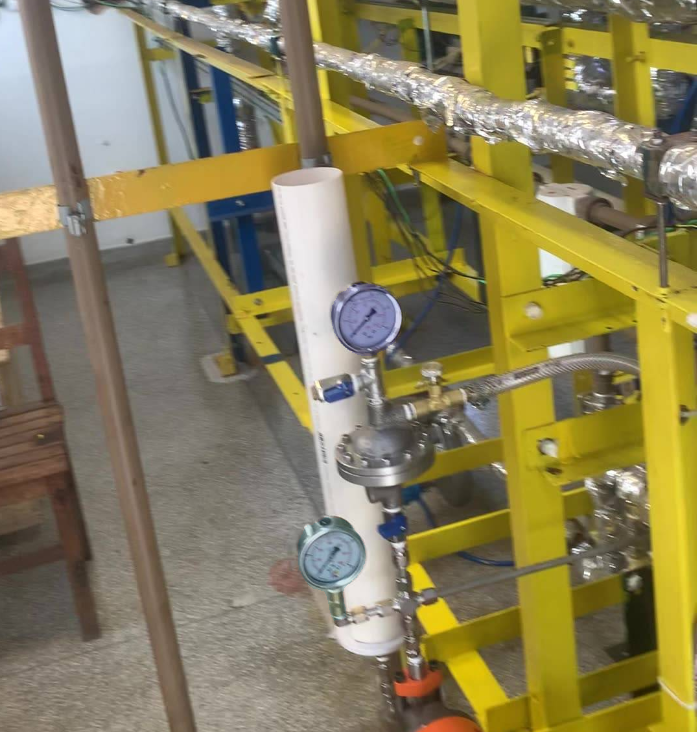}
    \caption{SAM2-assisted segmentation overlays.}
    \label{fig:real_sam2_seg1}
  \end{subfigure}
  \caption{Examples of real data annotations: (a) manual boxes curated in CVAT; (b) semi-automatic pixel masks using SAM2, later spot-checked by experts.}
  \label{fig:real_data_examples}
\end{figure}

We collected approximately 2,500 real images and video frames of industrial
manometers. These samples were annotated using a hybrid approach: manual bounding-box labeling with CVAT and semi-automatic segmentation with SAM2, as shown in Fig.~\ref{fig:real_data_examples}. A small portion of these annotations was further reviewed by experts to
ensure quality.

\subsection{Synthetic Data with BlenderProc}
\begin{figure}[H]
  \centering
  \begin{subfigure}[b]{0.45\linewidth}
    \centering
    \includegraphics[width=\linewidth]{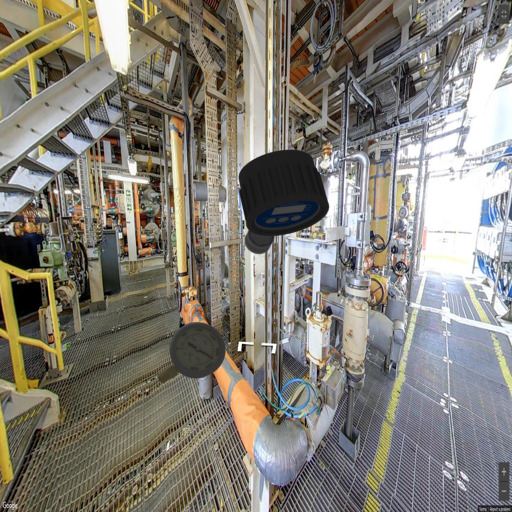}
    \caption{Manometer in a random background.}
    \label{fig:real_cvat_bbox2}
  \end{subfigure}\hfill
  \begin{subfigure}[b]{0.45\linewidth}
    \centering
    \includegraphics[width=\linewidth]{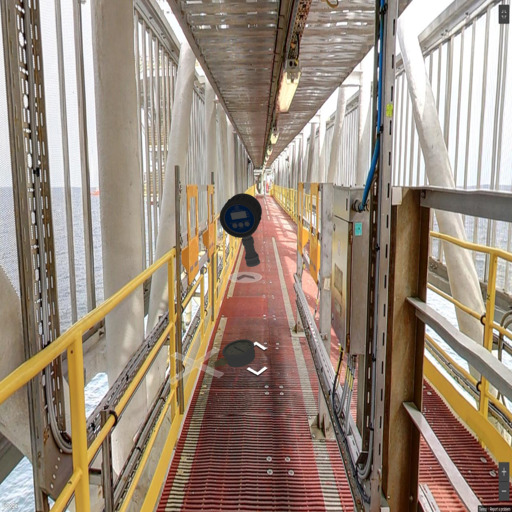}
    \caption{An example of multiple .blend files.}
    \label{fig:real_sam2_seg2}
  \end{subfigure}
  \caption{BlenderProc synthetic samples.}
  \label{fig:blenderproc}
\end{figure}

3D models of manometers were rendered with BlenderProc. To improve generalization and reduce overfitting, domain randomization was applied: random backgrounds (industrial textures and photos), lighting variations, camera pose sampling, and post-processing (noise, blur, chromatic aberration), with examples provided in Fig.~\ref{fig:blenderproc}. BlenderProc provides pixel-perfect labels, ensuring accurate
segmentation and bounding-box masks.

\subsection{AI-Extended Video Generation (ComfyUI + Cosmos-Predict2)}
\begin{figure}[H]
  \centering
  \begin{subfigure}[b]{0.48\linewidth}
    \centering
    \includegraphics[width=\linewidth]{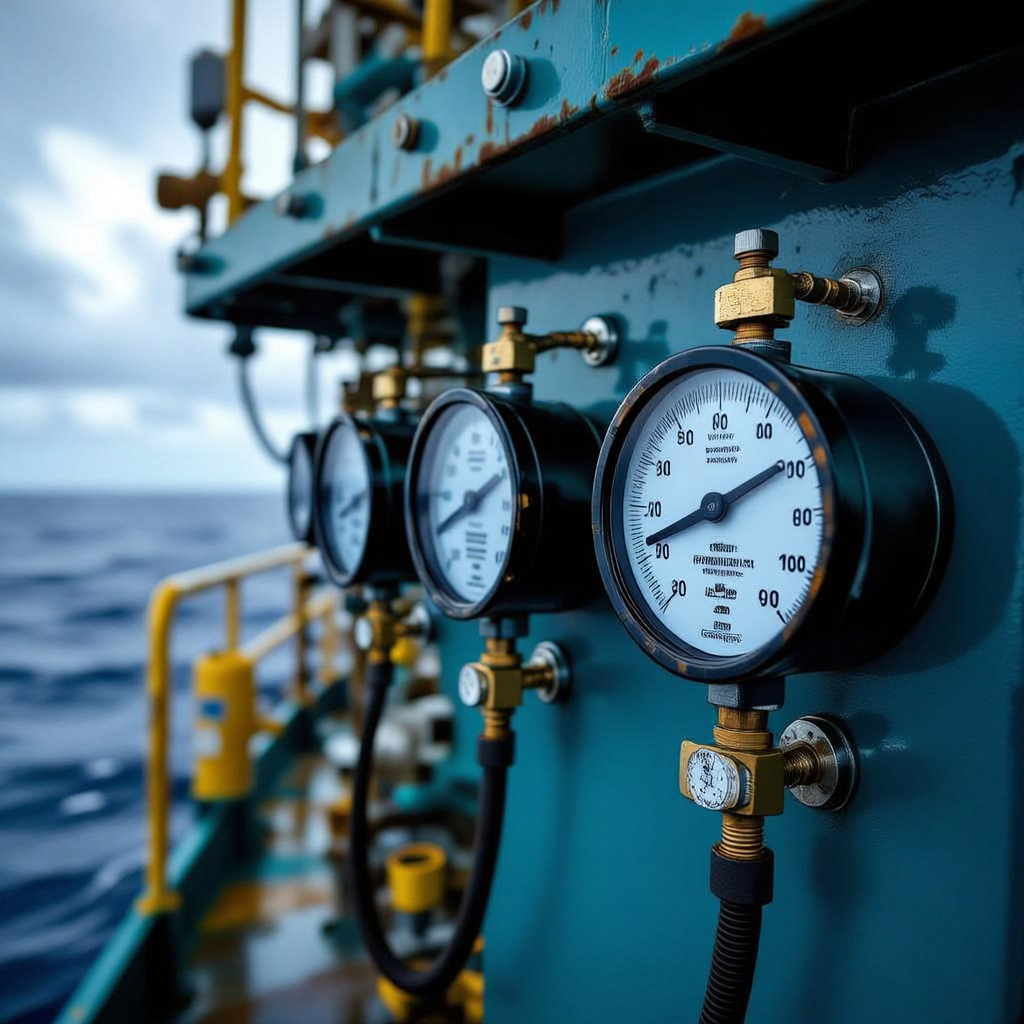}
    \caption{Relighting and viewpoint changes with motion blur.}
    \label{fig:real_cvat_bbox3}
  \end{subfigure}\hfill
  \begin{subfigure}[b]{0.48\linewidth}
    \centering
    \includegraphics[width=\linewidth]{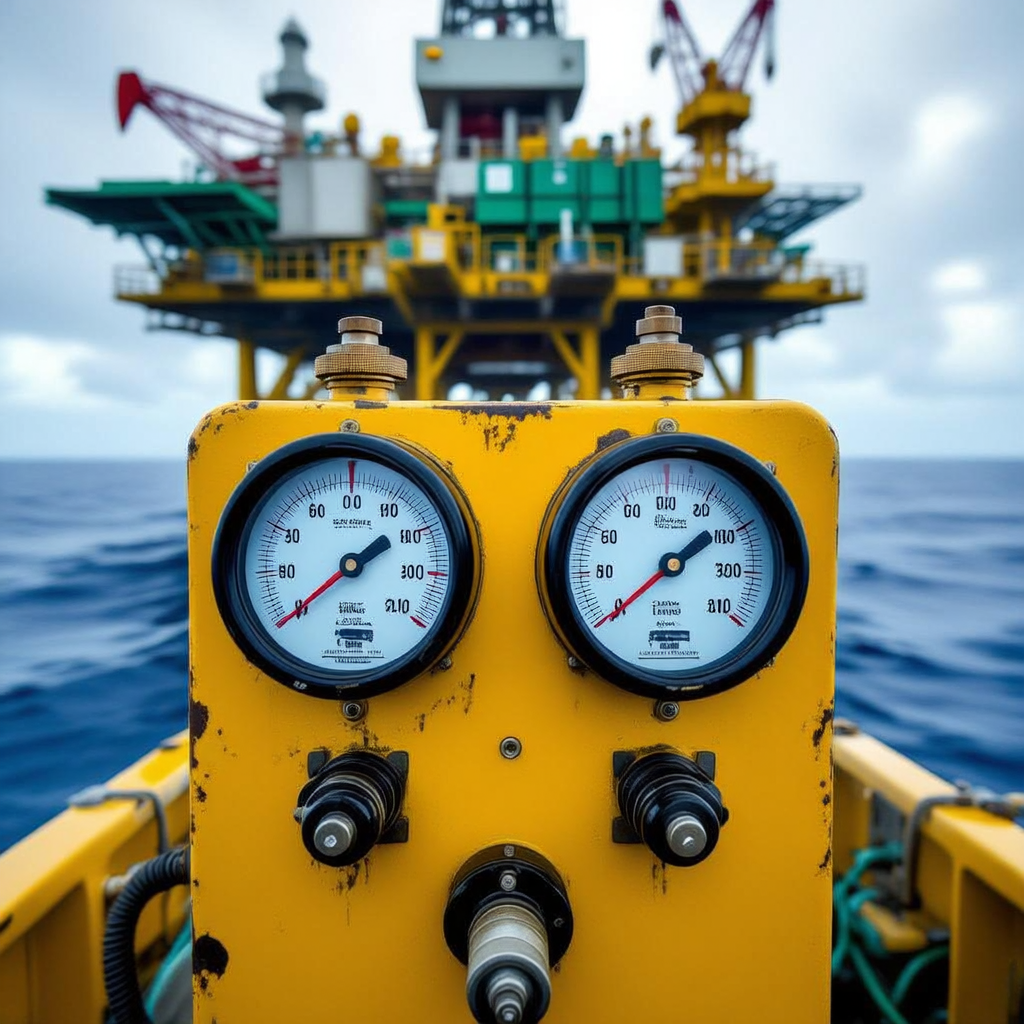}
    \caption{A frame from a short clip made with Cosmos.}
    \label{fig:real_sam2_seg3}
  \end{subfigure}
  \caption{AI-extended video frames produced via ComfyUI + NVIDIA Cosmos-Predict2.}
  \label{fig:AI-extended}
\end{figure}

To complement static synthetic renders, we integrated AI-driven video extension, illustrated in Fig.~\ref{fig:AI-extended}. Real short clips of manometers were expanded using ComfyUI
workflows with NVIDIA Cosmos-Predict2, enabling frame synthesis with
temporal consistency, relighting, viewpoint changes, and motion blur. These
pseudo-extended frames introduce temporal diversity (occlusion, reflections,
jitter) that is difficult to model with static rendering. Pseudo-labels were
propagated across frames using tracking and filtered with confidence
thresholds, with human-in-the-loop auditing on a subset to mitigate noise.

\subsection{Data Composition Strategy}
All data sources were merged into a unified dataset. To investigate the
impact of proportions, we defined scenarios with:
\begin{itemize}
    \item {Real only} (2,500 images)
    \item {Mixed} (real + synthetic in ratios 1:1, 1:3)
\end{itemize}

Within the synthetic portion, we tested different splits between BlenderProc
renders and AI-extended video frames, adopting 70\% BlenderProc and 30\%
video-based synthesis as baseline.

\subsection{Training and Evaluation}
\begin{figure}[!t]
  \centering
  \includegraphics[width=0.8\linewidth]{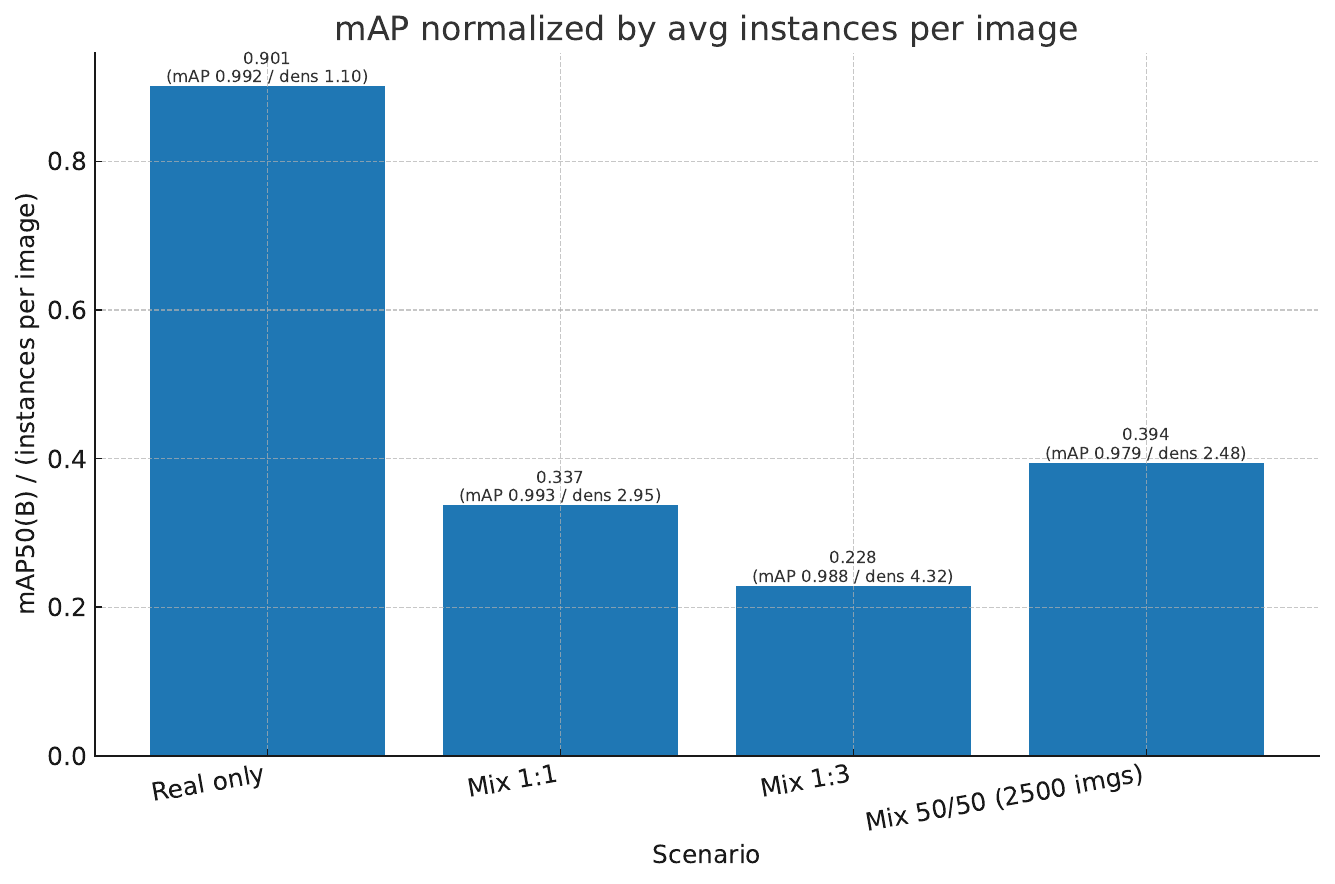}
  \caption{Normalized mAP (0.5–0.95) per dataset scenario, corrected by the average number of instances per image. By normalizing precision scores with respect to the number of object instances, this metric mitigates the effect of dataset imbalance and varying annotation densities}
  \label{fig:normalized}
\end{figure}

All scenarios were trained with the same YOLO-based architecture using
identical hyperparameters. Evaluation was performed exclusively on a held-out
set of real images. Metrics include mean Average Precision (mAP) at intersection-over-union (IoU) thresholds .5:.95, recall, Average Recall (AR), per-class Average Precision (AP), and precision–recall (PR) curves. We also report normalized metrics (mAP divided by average instances per image), shown in Fig.~\ref{fig:normalized}, and perform ablations on domain
randomization factors.

\subsection{Computational Setup}
\label{subsec:comp_setup}
 We ran the experiments on a local workstation with 2$\times$NVIDIA RTX A2000 12\,GB. The video synthesis process using ComfyUI required approximately 8 minutes to generate a 15-second video, which was subsequently decomposed at 20 frames per second, yielding an average of 300 frames per minute. In comparison, rendering 1000 images in Blender took approximately 13 minutes, corresponding to a throughput of 77 images per minute. Regarding the YOLO training times, Table~\ref{tab:treining time} summarizes the total training duration for each dataset configuration.

\begin{table}[h]
\centering
\caption{Total training duration}
\label{tab:treining time}
\begin{tabular}{lccc}
\hline
Scenario & Training Time  \\
\hline
    Real Only &             3h 18m 8s    \\
    Mix 1:1 &                1h 54m 45s \\
    Mix 1:3 &                3h 45m 39s \\
   Mix 0.5:0.5 &                1h 16m 47s \\
\end{tabular}
\end{table}

\section{RESULTS}\label{sec:results}

In this section we present the experimental protocol, justification for
synthetic data generation, proportionality choices between real and
synthetic samples, and the results obtained in different scenarios.

\subsection{Experimental Protocol}
Four training scenarios were defined: 
(i) only real images (2,500) and
(ii) a mix of real and synthetic images.

For mixed datasets, we tested ratios of 0.5:0.5, 1:1, and 1:3 (real:synthetic). 
All scenarios were trained with the same YOLO-based detection architecture
and evaluated on the same test split of real images.

\subsection{Justification}
The adoption of synthetic data is motivated by three factors: 
1) the cost and time of annotating real images are significantly higher than generating synthetic ones; 
2) domain randomization techniques such as random backgrounds, lighting and camera poses reduce overfitting and bridge the sim-to-real gap; 
3) prior work shows that mixing 5--20\% real data with synthetic data is often enough to reach competitive performance.

\subsection{Proportionality of Synthetic Data}
Considering the availability of 2,500 real photos, we defined proportionality
as follows:  
- {1:1} → 2,500 real + 2,500 synthetic  
- {1:3} → 2,500 real + 7,500 synthetic  
 
Inside the synthetic portion, we explored the impact of splitting between
BlenderProc-rendered images and AI-extended video frames. BlenderProc provides
proper labels and strong control over background, illumination, and pose, while
video-based synthesis injects temporal variation such as blur, occlusion, and
camera jitter. The baseline adopted was $70\%$ BlenderProc and
$30\%$ video-based frames, which balances labeling accuracy and temporal
diversity.

\subsection{Results}
Table~\ref{tab:datasets} shows dataset composition and Table~\ref{tab:performance} reports object detection performance.

\begin{table}[h]
\centering
\caption{Dataset composition (examples).}
\label{tab:datasets}
\begin{tabular}{lccc}
\hline
Scenario & Real Images & Synthetic Images & Total \\
\hline
Real only & 2500 & 0 & 2500 \\
Mix 1:1 & 2500 & 2500 & 5000 \\
Mix 1:3 & 2500 & 7500 & 10000 \\
Mix 0.5:0.5 & 1250 & 1250 & 2500 \\
\hline
\end{tabular}
\end{table}

\begin{table}[h]
\centering
\caption{Performance comparison (updated with 0.5:0.5).}
\label{tab:performance}
\begin{tabular}{lccc}
\hline
Scenario     & mAP@[.5:.95] & Recall & AR \\
\hline
Real only    & 0.936 & 0.969 & 0.969 \\
Mix 1:1      & 0.962 & 0.972 & 0.972 \\
Mix 1:3      & 0.928 & 0.948 & 0.948 \\
Mix 0.5:0.5  & 0.868 & 0.933 & 0.933 \\
\hline
\end{tabular}
\end{table}

\begin{figure}[!t]
  \centering
  \includegraphics[width=1\linewidth]{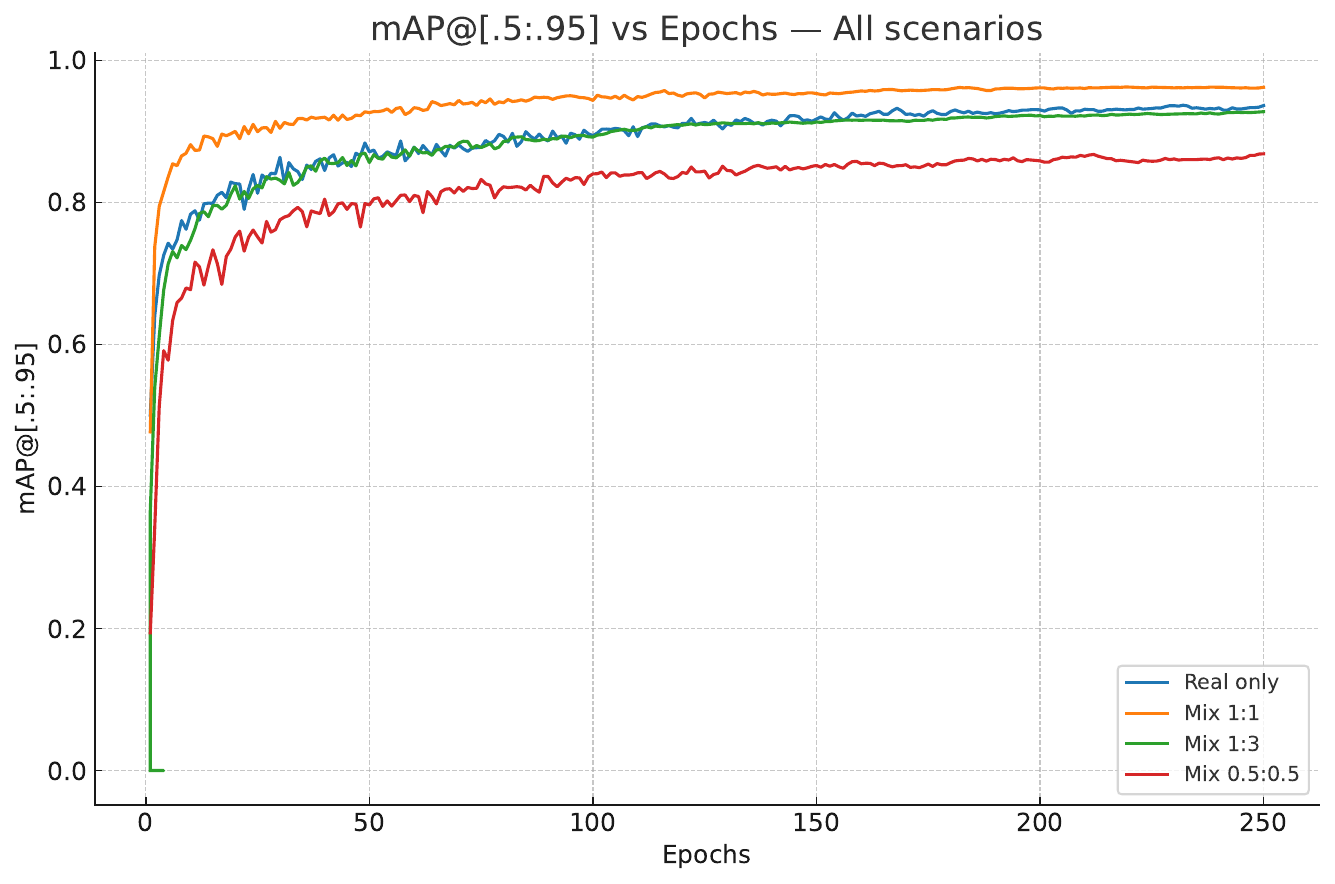}
  \caption[w]{ These results indicate that adding synthetic data is beneficial, but excessive synthetic proportion (e.g., 1:3) does not yield further gains over Real-only and incurs additional compute cost.}

  \label{fig:map_scenario}
\end{figure}

These results indicate that mixed 1:1 dataset outperforms the real-only baseline. In our experiments, the 1:1 mix achieved the highest mAP50-95, while the 1:3 mix still surpassed the real-only model but trailed the 1:1 configuration, as depicted in the training evolution in Fig.~\ref{fig:map_scenario}. These findings corroborate the broader consensus that blending a modest amount of real data with a larger volume of synthetic data can guide the detector toward realistic features while benefiting from the diversity and volume of synthetic samples.

\subsection{Additional Analyses}

\begin{table}[h]
\centering
\caption{Number of instances in each scenario}
\label{tab:instances}
\begin{tabular}{lccc}
\hline
Scenario     &Instances \\
\hline
Real only    & 2753  \\
Mix 1:1      & 13560  \\
Mix 1:3      & 43242  \\
Mix 0.5:0.5  & 6207  \\
\hline
\end{tabular}
\end{table}

\begin{figure}[!htbp]
  \centering
  \includegraphics[width=1\linewidth]{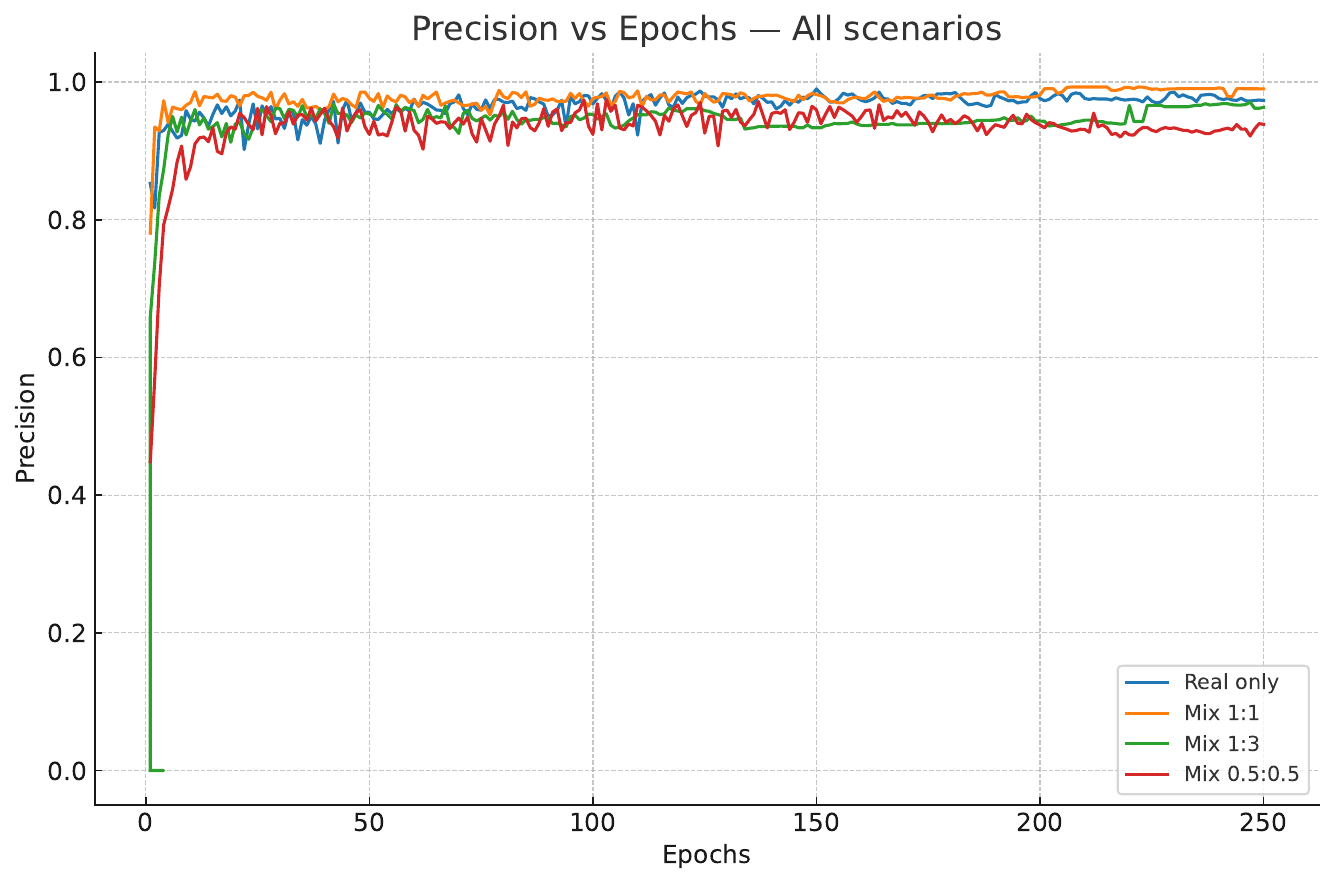}
  \caption{Precision over training epochs for four training regimes. The synthetic-augmented configuration reaches high precision within the first epochs and remains near saturation. Synthetic data can accelerate the reduction of false positives.}
  \label{fig:plot_precision_curves}
\end{figure}

\begin{figure}[!htbp]
  \centering
  \includegraphics[width=1\linewidth]{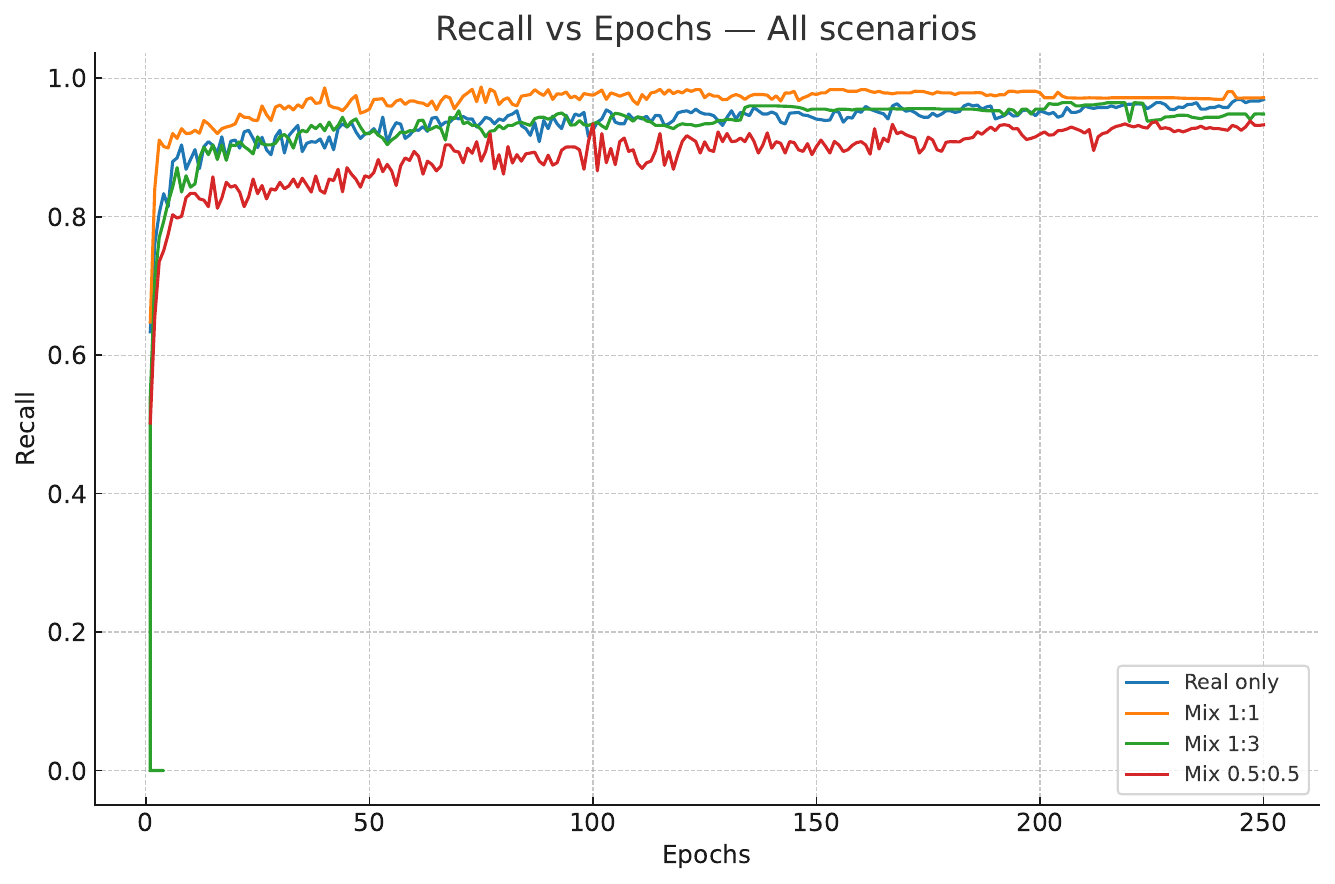}
   \caption{Recall across training epochs for four training regimes. Mix 1:1 provides the most consistent recall gains. Real-only and Mix 1:3 display similar recall trajectories, whereas Mix 0.5:0.5 is lower yet notably better than a naive halving of the real dataset would suggest, evidencing the compensatory effect of synthetic data.}
  \label{fig:plot_recall_curves}
\end{figure}

\begin{figure}[!htbp]
  \centering
  \includegraphics[width=1\linewidth]{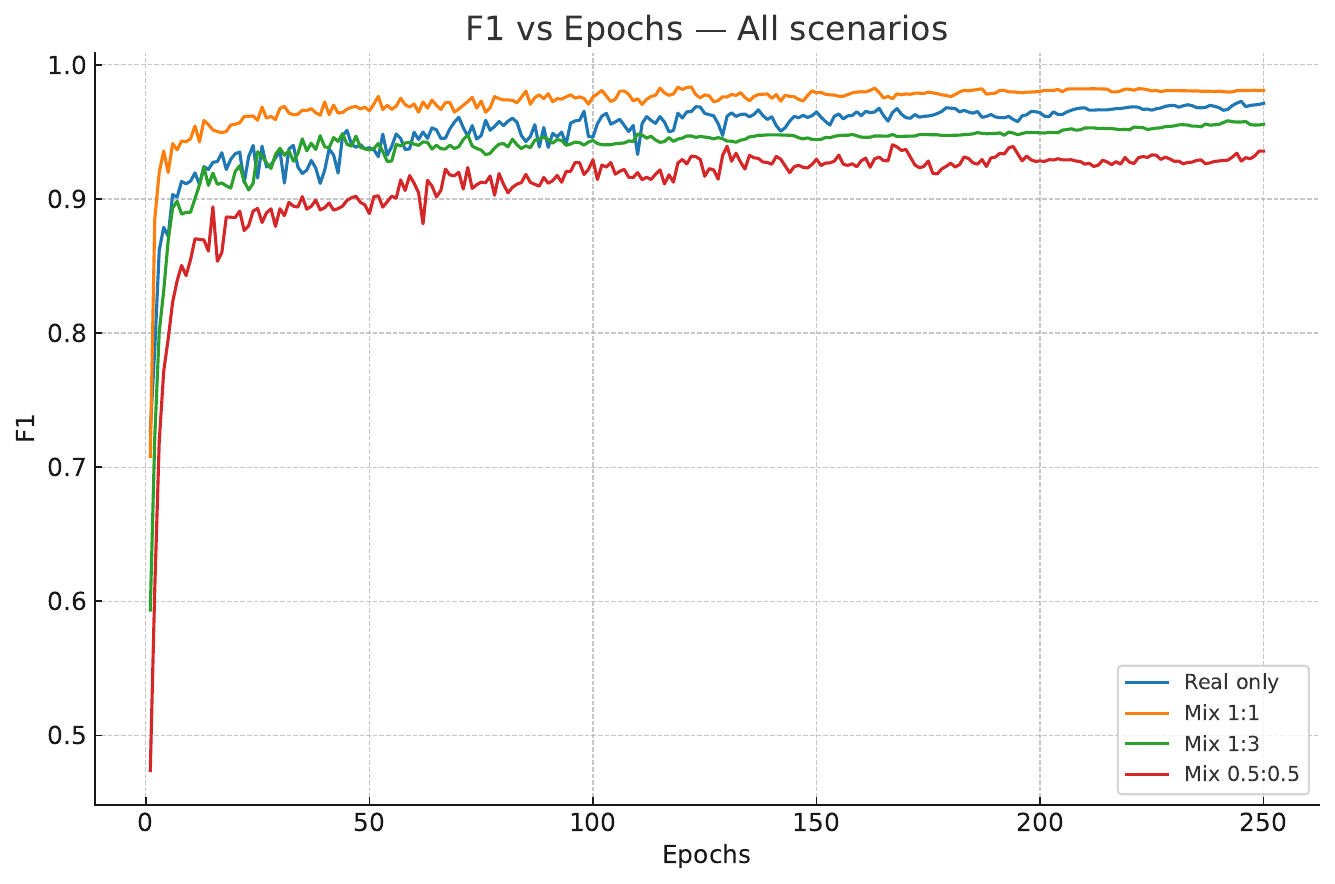}
  \caption{F1-score over epochs for four training regimes. The curves mirror the recall advantage of Mix 1:1 while preserving high precision. Real-only and Mix 1:3 are nearly tied, and Mix 0.5:0.5 trails but benefits from synthetic augmentation given the reduced real data.}

  \label{fig:plot_f1_curves}
\end{figure}

\begin{figure}[!htbp]
  \centering
  \includegraphics[width=1\linewidth]{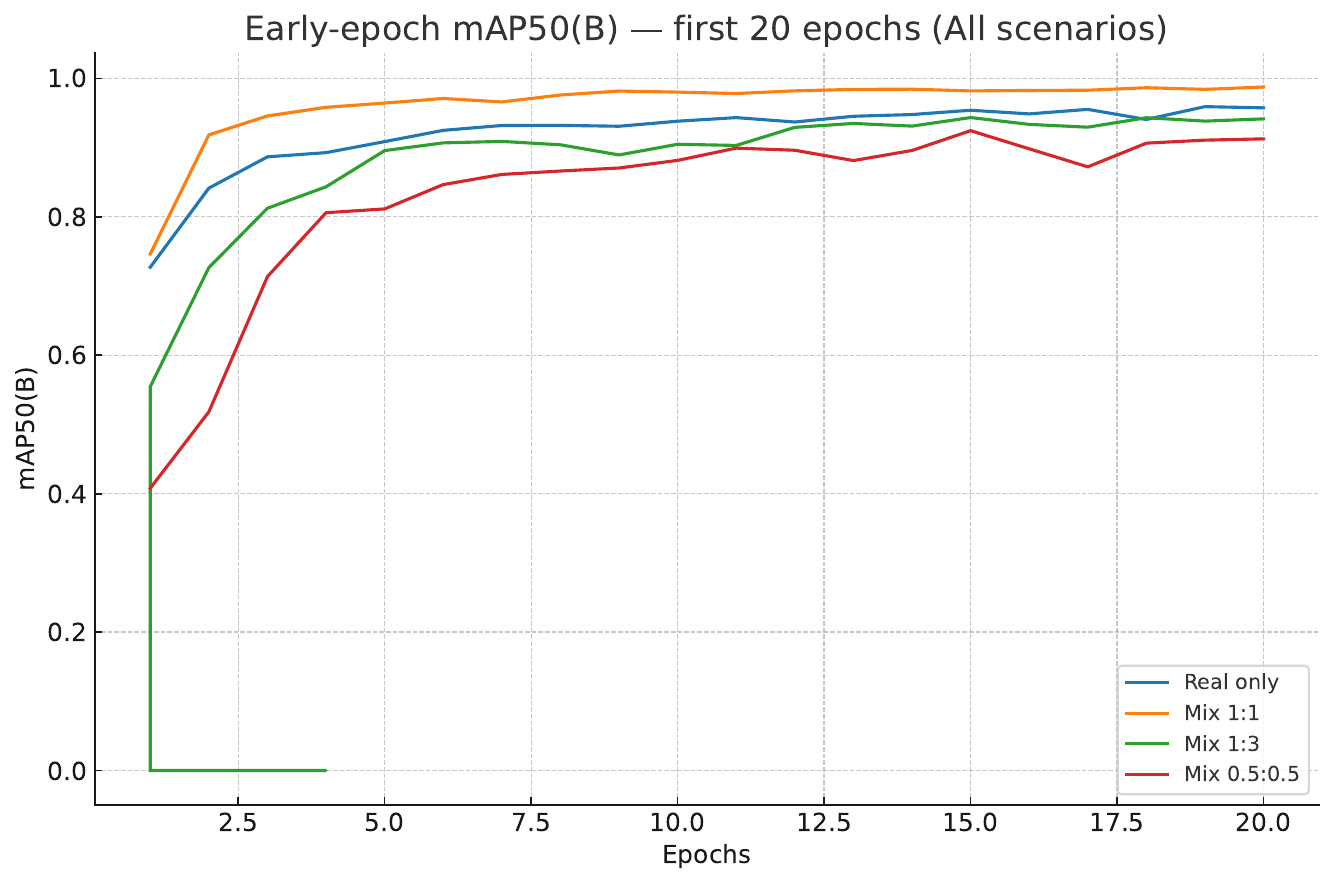}
  \caption{Early-epoch mAP50(B) (first 20 epochs) comparing four training regimes.}
  \label{fig:learning_curves_custom}
\end{figure}

\FloatBarrier

Figures~\ref{fig:plot_precision_curves}--\ref{fig:plot_f1_curves} showcase the training dynamics of \emph{Precision}, \emph{Recall}, and \emph{F1-score} across the four regimes (Real only, Mix~1:1, Mix~1:3, and Mix~0.5:0.5), enabling a side-by-side inspection of convergence behavior. A consistent pattern is that the Mix~1:1 configuration accelerates early learning relative to the Real-only baseline, reflected in higher early-epoch averages. In contrast, Mix~1:3 tends to underperform in the same interval and later converges to results that are practically tied with Real only. The Mix~0.5:0.5 setting exhibits the slowest convergence and lowest asymptotic accuracy; nevertheless, despite using only half the real images, synthetic augmentation prevents a sharp degradation, indicating that synthetic data can partially compensate for small or imperfect real dataset. Taken together, these analyses suggest that synthetic data can significantly improve early optimization and stability, especially in Recall and the resulting F1 score, provided that the synthesis pipeline is properly configured; otherwise, suboptimal choices in synthetic generation may negate the expected benefits.

\iffalse
\subsubsection{Precision–Recall Curves}
We provide per-scenario PR curves to complement mAP and highlight the precision–recall trade-off. Mixed datasets tend to shift the curve upward/left, indicating higher precision at comparable recall levels, consistent with prior observations on domain randomization and mixed supervision~\cite{manufacturing2025,neurolabs2023,powerSynthetic2019}.

\fi

\section{CONCLUSIONS}\label{sec:conc}
This paper showed that combining synthetic and real images enhances detection performance and training stability in industrial settings. A 1:1 mixture achieves the highest overall accuracy, while real-only and 1:3 mixes yield similar results, indicating diminishing returns when synthetic data dominates. Even with only half the real images, synthetic augmentation mitigates performance drops, highlighting its value for small or imperfect datasets.

Limitations include evaluation on a single asset class and camera setup, focus on 2D detection only, potential temporal artifacts from video synthesized frames, and no benchmarking on embedded platforms. Future work will expand to multiple asset types, richer tasks (instance segmentation, 6D pose, needle angle/OCR), multimodal sensing, cross-domain adaptation, and on-robot validation with embedded hardware, incorporating active learning and quality/uncertainty monitoring. This approach can support more robust, cost-effective industrial perception systems.

\section{DATA AVAILABILITY}
The data that support the findings of this study are available in Figshare at \hyperlink{https://doi.org/10.6084/m9.figshare.29936768.v1}{doi.org/10.6084/m9.figshare.29936768.v1}, reference number 29936768 \cite{SyntheticDataManometryInterpretation}.

% \section{ACKNOWLEDGMENTS}
% This work was carried out with the support of Petrobras, using resources from the R\&D clause of the ANP, in partnership with the University of São Paulo (USP) and the intervening foundation Fundação de Apoio à Física e à Química (FAFQ), under Cooperation Agreement No. 2023/00013-7 and 2023/00016-6.

%%%%%%%%%%%%%%%%%%%%%%%%%%%%%%%%%%%%%%%%%%%%%%%%%%%%%%%%%%%%%%%%%%%%%%%%%%%%%%%%

\bibliographystyle{IEEEtran}
\bibliography{ref}

\end{document}